\documentclass[a4paper]{article} 
\usepackage{times}
\usepackage[margin=2.5cm,top=2cm]{geometry}

\usepackage[square,sort]{natbib}
\renewcommand{\cite}[1]{\citep{#1}}

\usepackage[utf8]{inputenc} 
\usepackage[T1]{fontenc}    
\usepackage{booktabs}       
\usepackage{nicefrac}       
\usepackage{microtype}      

\usepackage{graphicx}
\usepackage{amsmath}
\usepackage{amssymb}
\usepackage{array}
\usepackage{hyperref}       
\usepackage{url}            

\graphicspath{{../graphics/}}
\usepackage{multirow}

\newcommand{\Fig}[1]{Figure~\ref{#1}}
\newcommand{\fig}[1]{Fig.~\ref{#1}}
\newcommand{\Tab}[1]{Table~\ref{#1}}
\newcommand{\tab}[1]{Tab.~\ref{#1}}

\newcommand{\eqn}[1]{Eq.~\eqref{#1}} 
\newcommand{\eqnp}[1]{(Eq.~\ref{#1})} 

\newcommand{\ie}{i.\,e.~}
\newcommand{\eg}{e.\,g.~}
\newcommand{\wrt}{w.\,r.\,t.~}
\newcommand{\Real}{\ensuremath{\mathbb R}}        
\newcommand{\sigm}{\ensuremath{\text{sigm}}}                
\newcommand\Tstrut{\rule{0pt}{2.6ex}}
\DeclareMathOperator*{\argmin}{arg\,min}

\newcommand{\method}{EQL} 
\newcommand{\typ}{\ensuremath{I}}             
\newcommand{\layer}[1]{{(#1)}}           
\renewcommand{\l}{{\layer{l}}}           
\newcommand{\lm}{{\layer{l-1}}}           

\usepackage[disable]{todonotes}
\begin{document}
\title{Extrapolation and learning equations}
\author{%
  Georg Martius \& Christoph H. Lampert\\
  IST Austria\\
  Am Campus 1, 3400 Klosterneuburg, Austria\\
  \texttt{\{gmartius,chl\}@ist.ac.at}
}
\maketitle

\begin{abstract}
In classical machine learning, regression is
treated as a black box process of identifying a
suitable function from a hypothesis set
without attempting to gain insight into the mechanism connecting inputs and  outputs.
In the natural sciences, however, finding an interpretable function for a phenomenon
is the prime goal as it allows to understand and  generalize results.
This paper proposes a novel type of function learning
network, called equation learner (EQL),
that can learn analytical expressions and is able to extrapolate to unseen domains.
It is implemented as an end-to-end differentiable feed-forward network and allows for
 efficient gradient based training.
Due to sparsity regularization concise interpretable expressions can be obtained.
Often the true underlying source expression is identified.
\end{abstract}

\section{Introduction}\label{sec:intro}
The quality of a model is typically measured by
its ability to generalize from a training set to previously unseen
data from the same distribution.
In regression tasks generalization essentially boils down to interpolation
 if the training data is sufficiently dense.
As long as models are selected correctly, \ie
 in a way to not overfit the data, the regression problem is well understood
and can -- at least conceptually -- be considered solved. %
However, when working with data from real-world devices, \eg
controlling a robotic arm, interpolation might not be sufficient.
It could happen that future data lies outside of the training
domain, \eg when the arm is temporarily operated outside of its
specifications.
For the sake of robustness and safety it is desirable in such a case
to have a regression model that continues to make good predictions,
or at least does not fail catastrophically.
This setting, which we call \emph{extrapolation generalization}, is
the topic of the present paper.

We are particularly interested in regression tasks for systems that
can be described by real-valued analytic expression, \eg mechanical systems
such as a pendulum or a robotic arm.
These are typically governed by a highly nonlinear function but it is
nevertheless possible, in principle, to infer their behavior
on an extrapolation domain from their behavior elsewhere.
%
We make two main contributions: 1) a new type of
 network that can learn analytical expressions and is able to extrapolate to unseen domains
 and 2) a model selection
 strategy tailored to the extrapolation setting.

The following section describes the setting of regression and extrapolation.
Afterwards we introduce our method and discuss the architecture,
 its training, and its relation to prior art. 
We present our results in the Section \emph{Experimental evaluation}  
  and close with conclusions.

\section{Regression and extrapolation}\label{sec:setting}
We consider a multivariate regression problem with a training
set $\{(x_1,y_1),\dots,(x_N,y_N)\}$ with $x \in \Real^n$, $y\in \Real^m$. 
Because our main interest lies on extrapolation in the context of
learning the dynamics of physical systems we assume
 the data originates from an unknown analytical function (or system
of functions),
$\phi:\Real^n\to\Real^m$ with additive zero-mean noise, $\xi$, \ie
$y=\phi(x)+\xi$ and $\mathbb{E}\xi=0$.
The function $\phi$ may, for instance, reflect a system of
ordinary differential equations that govern the movements of
a robot arm or the like.
The general task is to learn a function $\psi:\Real^n\to\Real^m$
that approximates the true functional relation as well as possible
in the squared loss sense, \ie achieves minimal expected error $\mathbb{E}\|\psi(x) - \phi(x)\|^2$.
In practice, we only have particular examples of the function values
available and measure the quality of predicting in terms of the
empirical error on training or test data $D$, 
\begin{align}
  E(D)&=\frac{1}{N}\sum^{N}_{i=1}\|\psi(x_i) - y_i\|^2\,. \label{eqn:error}
\end{align}
If training and test data are sampled from the same distribution
then we speak about an \emph{interpolation} problem.
In the \emph{extrapolation} setting the training data is assumed to
cover only a limited range of the data domain.
In the example of the robot arm, for instance, the training may
be restricted to a certain joint angle range or maximal velocity.
For testing we want to make predictions about the unseen
domains, \eg for higher velocities.
As usual, we split the data that is available at training time
into a part for model training 
and a part for validation or model selection. 

\section{Learning a network for function extrapolation}\label{sec:method}
The main model we propose is a multi-layered feed-forward network
with computational units specifically designed for the extrapolation
regression tasks.
For an $L$-layer network, there are $L-1$ hidden layers, each consisting
of a linear mapping followed by non-linear transformations.
For simplicity of notation, we explain the network as if each hidden
layer had the same structure ($k'$ inputs, $k$ outputs).
In practice, each layer can be designed independently of the others,
of course, as long as input/output dimensions match.

The linear mapping at level $l$ maps the $k'$-dimensional input $y^{\lm}$ to the $d$-dimensional
 intermediate representation $z$ given by
\begin{align}
  z^\l &= W^\l y^\lm + b^\l,
\end{align}
where $y^\lm$ is the output of the previous layer, with the convention $y^{(0)}=x$.
The weight matrix $W^\l\in \Real^{d \times k'}$ and the bias vector $b^\l\in\Real^{d}$
are free parameters that are learned during training.
The non-linear transformation contains $u$ \emph{unary units},
$f_i:\Real\to\Real$, for $i=1,\dots,u$, and $v$ \emph{binary units},
$g_j:\Real\times\Real\to\Real$ for $j=1,\dots,v$. Their outputs are
concatenated to form the layer output
\begin{align}
  y^\l &:= \Big(f_1(z^\l_1),f_2(z^\l_2),\dots,f_{u}(z^\l_{u}),\nonumber\\
  & \qquad g_{1}(z^\l_{u+1},z^\l_{u+2}),\dots,g_{v}(z^\l_{u+2v-1},z^\l_{u+2v}) \Big)\,.
\end{align}

\begin{figure}
  \centering
  \includegraphics[width=0.6\linewidth]{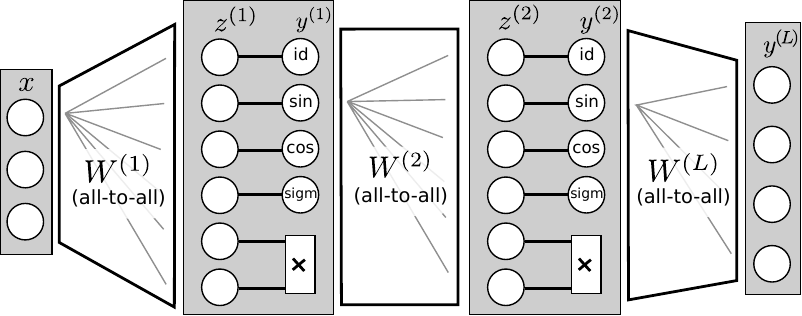}
  \caption{Network architecture of the proposed Equation Learner (\method) for 3 layers ($L=3$) and one neuron per type ($u=4,v=1$).}
  \label{fig:network}
\end{figure}

In total, the nonlinear stage has $k = u + v$ outputs and  $d = u + 2 v$ inputs.
The unary units, $f_1,\dots,f_u$ receive the respective component, $z_1,\dots,z_u$
as inputs, and each unit may be one of the following base functions as specified
in a fixed type parameter $\typ_i\in\{0,1,2,3\}$
\begin{align}
  f_i(z_i) &:=
  \begin{cases}
    z_i & \text{ if } \typ_i=0,\\
    \sin(z_i) & \text{ if } \typ_i=1,\\
    \cos(z_i) & \text{ if } \typ_i=2,\\
    \sigm(z_i) & \text{ if } \typ_i=3,
  \end{cases}&\text{ for } i=1,\dots,u,
\end{align}
where $\sigm(z)=\frac{1}{1+e^{-z}}$ is the standard sigmoid function.
The binary units, $g_1,\dots,g_v$ receive the remaining component, $z_{u+1},\dots,z_{u+2v}$,
as input in pairs of two. They are \emph{multiplication units} that compute the product of
their two input values:
\begin{align}
  g_j(z_{u+2j-1}, z_{u+2j}) &:= z_{u+2j-1} \cdot z_{u+2j}&\text{ for }j=1,\dots,v.
\end{align}
Finally, the $L$-th and last layer computes the regression values by a linear read-out
\begin{align}
  y^{\layer{L}} &:= W^{\layer{L}} y^{\layer{L-1}} + b^{\layer{L}}.
\end{align}
The architecture is depicted in \fig{fig:network}.
We call the new architecture Equation Learner (\method{}) and
 denote the function it defines by $\psi$.

\subsection{Discussion of the architecture}
The proposed network architecture differs in two main aspects
from typical feed-forward networks: the existence of multiplication
units and the possibility of \emph{sine} and \emph{cosine} as nonlinearities for the unary units.
Both design choices are motivated by our objective of learning
a system of equations that govern a physical
system and can extrapolate to new parts of the input space.

\emph{Sigmoid} nonlinearities are the canonical choice of
\emph{activation function} for \emph{artificial neural networks} (ANN)
 and proved to be successful.
In fact, we include sigmoids in our architecture, making it a super class of ANNs.
However, they were typically disabled by the training procedure
corresponding to their absence in the considered physical equations.
Other, predominantly local nonlinearities, in particular
\emph{radial basis functions}~\cite{broomhead1988radial} we
do not include, since one cannot expect them to
extrapolate at all.
Further nonlinearities, such as \emph{(square) roots} and
\emph{logarithms}, could in principle be useful for
learning physical equations, but they pose problems because
their domains of definition is restricted to positive
inputs.
We leave the task of incorporating them in a principled way to future work.

The ability to multiply two values is a second crucial component
of our network architecture.
Again, it is inspired by the typical form of physical equations,
where multiplication of components is arguably second common
basic operation after addition (which the linear layers can perform).
Multiplication was introduced into neural networks long
ago as product-units~\cite{DurbinRumelhart1989:ProductUnits} and Pi-Sigma-unit~\cite{ShinGhosh1991:pi-sigma}. The product-units have large fan-in that compute products over
all their inputs, potentiated by the respective weights.
The result is typically the behavior of a high order polynomial,
which are powerful function approximators, but rarely occur in
physical equations. Polynomials are also known to require careful
fine-tuning in order not to overfit, which makes them a risky
choice for the purpose of extrapolation.
The Pi-Sigma units are multiplication units with a fixed number of factors
 and our multiplication units are a special for 2 factors.
We find that multiplying just two values at a time is well adjusted
to the task we aim at, as it allows to control the maximal degree of
the learned polynomial by the depth of the network.

Finally, each layer of the network contains unary units that act
as \emph{identity} maps, which in particular gives the network
the option to learn functions with smaller number of nonlinearities
than the total network depths.

\subsection{Network training}
The \method{}  is fully differentiable in its free parameters
$\theta=\{W^{(1)},\dots,W^{(L)},b^{(1)},\dots,b^{(L)}\}$,
which allows us to train it in an end-to-end fashion using back-propagation.
We adopt a Lasso-like objective~\cite{tibshirani1996regression},
\begin{align}
  L(D)&=\frac{1}{N}\sum^{|D|}_{i=1}\|\psi(x_i) - y_i\|^2 + \lambda \sum_{l=1}^L\big|W^\l\big|_1\,,\label{eqn:loss}
\end{align}
that is, a linear combination of $L_2$ loss and $L_1$ regularization,
and apply a stochastic gradient descent algorithm with mini-batches and Adam~\cite{KingmaBa2015:Adam}
 for calculating the updates:
\begin{align}
  \theta_{t+1} &= \theta_{t} + \text{Adam}\left(\frac{\partial L(D_{(t)})}{\partial \theta}, \alpha\right),
\end{align}
where $D_{(t)}$ denotes the current mini-batch and $\alpha$ is the stepsize parameter.
The choice of Adam is not critical and standard stochastic gradient descent also works.
In all numerical experiments we use $\alpha=0.001$ and a mini-batch size of 20.

The role of the $L_1$ regularization is to encourage networks with sparse connections,
matching the intuition that a typical formula describing a physical system contains
only a small number of terms, each operating only on a few variables.
However, in a non-convex setting where local minima are likely to occur, this type
of regularization can have an undesirable side-effect: during the course of the
optimization the weights hardly ever change their sign.
The reason is that the regularization leads to a constant rate of weight decay
whereas the counteracting derivative with respect to the square loss is proportional to the backpropagated
error signal and the input to the unit.
The latter contributions are often smaller along paths with small weights,
 such that many weights go to zero and stay there.
Additionally, any non-zero regularization term causes the learned weights to reflect a
trade-off between minimizing the loss and the regularizer.
Although, this can lead to improved generalization,
 it also results in a systematic underestimation of the function values.

Therefore, we follow a hybrid regularization strategy: at the beginning of the training
procedure ($t<t_1$) we use no regularization ($\lambda=0$), such that parameters
can vary freely and reach reasonable starting points.
Afterwards, we switch on the regularization by setting $\lambda$ to a nonzero value,
which has the effect that a sparse network structure emerges.
Finally, for the last steps of the training ($t>t_2$) we disable $L_1$ regularization ($\lambda=0$)
 but enforce the same $L_0$ norm of the weights.
This is achieved by keeping all weights $w\in W^{1\dots L}$ that are close to 0 at 0,
 \ie if $|w|<0.001$ then $w=0$ during the remaining epochs.
This ensures that the learned model finds not only a function of the right parametric
form, but also fits the observed values as closely as possible.
We observed that the exact choice of breakpoints $t_1$ and $t_2$ is not critical. In practice, we
use $t_1 = \frac{1}{4} T$ and $t_2=\frac{19}{20} T$, where $T$ is total number of
update steps. $T$ was selected large enough to ensure convergence.
Note, that convergence to a sparse structure is important here, so
 early stopping will be disadvantageous.

\subsection{Model selection for extrapolation}\label{sec:modelsel}
\method{} networks have a number of hyper-parameters, \eg the number of layers,
 the number of units and the regularization constant.
Unfortunately, standard techniques for model selection, such as evaluation on a hold-out set or
cross-validation, will not be optimal for our purpose, since they rely
on interpolation quality.
In order to extrapolate the network has to find the ``right'' formula.
But how can we tell?
Using Occams razor principle: the simplest formula is most likely the right one.
Intuitively, if we have the choice between  $cos(x)$ and its truncated power series approximation $1-x^2/2 + x^4/24$, the first one is preferred.
We use the number of active hidden units
in the network as a proxy for the complexity of the formula, see Appendix A1 
for details.
One could also think of differentiating between the unit types.
In any case, this argumentation is only correct if the model explains the data well, \ie
 it has a low validation error.
So we have a dual objective to minimize, which we solve by ranking the instances
 \wrt validation error and sparsity
 and select the one with the smallest $L_2$ norm (in rank-space), see \eqn{eqn:model:sel}.

Furthermore, the optimization process may only find a local optimum of the training objective,
which depends on the initialization of the parameters.
We use independent runs to quantify expected performance deviations.

\subsection{Related work}%
In the field of machine learning, regression is often
treated as a black box process of identifying a
suitable real-valued function from a hypothesis set, \eg
a reproducing kernel Hilbert space for Gaussian Processes
Regression (GPR)~\cite{williams2006gaussian} or Support
Vector Regression (SVR)~\cite{smola2004tutorial},
or a multi-layer network of suitable expressive
power~\cite{specht1991general}.
The goal is to find a prediction function that leads to a small
expected error on future data, not necessarily to gain insight
into the mechanism of how the output values derive from the inputs.
The goal of finding an interpretable function is rather common in
the natural sciences, such as biology, where high noise levels and
strong inter-system variability often make it important to rely on
external prior knowledge, and finding a ``biologically plausible''
model is often preferable over finding one that makes the highest
prediction accuracy.
As a consequence, model classes are often highly constrained,
\eg allowing only for sparse linear models.

The task of learning a true, nonlinear, functional dependence
from observing a physical system, has received little attention
in the machine learning literature so far, but forms the basis
of the field of \emph{system identification}.
There, typically the functional form of the system is known and only
 the parameters have to be identified.
Another approach is to model the time evolution with autoregressive models
 or higher order convolution integrals (Volterra series)
 but learning analytic formulas is not common.

\emph{Causal learning} is an area of recent research that aims at
identifying a causal relation between multiple observables, 
which are typically the result of a physical process.
Classically, this tasks reduces to finding a minimal graphical
model based only on tests of conditional independence~\cite{Pearl2000}.
Although very successful in some fields, this classical approach
only provides a factorization of the problem, separating causes
and effects, but it leaves the exact functional dependency unexplained. 
Recent extensions of causal learning can take a functional view,
but typically do not constrain the regression functions to
physically plausible ones, but rather constrain the noise
distributions~\cite{PetersMJS2014}. 
The topic of learning a regression function with emphasis on
\emph{extrapolation} performance has not been studied much
in the literature so far. Existing work on time series
prediction deals with extrapolation in the temporal domain,
\ie predict the next value(s)~\cite{wiener1949extrapolation}. By our nomenclature,
this is typically rather an interpolation task, when the
prediction is based on the behaviour of the series at
earlier time steps but with similar value distribution~\cite{muller1997predicting,gyorfi2013nonparametric}.
Extrapolating in the data domain implies that the data
distribution at prediction time will differ from the data
distribution at training time. This is traditionally called
the \emph{domain adaptation} setting. In particular, since we
assume a common labeling function, our setting would fall
under the \emph{covariate shift} setting~\cite{quionero2009dataset}.
Unfortunately, this connection is not particularly useful for our
problem. As domain adaptation typically does not make additional
assumptions about how the data distribution may change, existing
methods need access to some unlabeled data from the test
distribution already at training time~\cite{ben2010theory}. In  
our setting this is not possible to obtain.

On the technical level, \method{} networks are an instance of
general feed-forward networks for function approximation~\cite{bishop1995neural}.
In contrast to recent trends towards \emph{deep learning}~\cite{bengio2009learning,bengio2013representation},
our goal is not to learn any data representation,
 but to learn a function which compactly represents
the input-output relation and generalizes between different regions
 of the data space, like a physical formula.
Structurally, \method{} networks resemble
\emph{sum-product networks (SPNs)}~\cite{PoonDomingos2011:sum-product-networks} and \emph{Pi-Sigma networks (PSNs)}~\cite{ShinGhosh1991:pi-sigma},
in the sense that both are based on directed acyclic
graphs with computational units that allows for summation
and multiplication.
Otherwise, SPNs are different as they act as efficient
alternative to probabilistic graphical models for representing
probability distributions, whereas \method{} networks are meant for
the classical task of function approximation.
In PSNs each output needs to be passed through multiplicative units, whereas in \method{} multiplication is optional.

Finding equations for observations is also known as symbolic regression
where a search is performed in a certain function space, typically done with evolutionary computation.
With these techniques it is possible to discover physical laws such as invariants and conserved quantities~\cite{SchmidtLipson2009:learnnaturallaws}.
Unfortunately, the computational complexity/search time explodes for larger expressions and high-dimensional problems.
We attempt to circumvent this by modeling it as a gradient based optimization problem.
Related to symbolic regression is finding mathematical identities for instance to find computationally more efficient expressions. In \cite{ZarembaFergus2014:LearnMathIdentities} this was done using machine learning to overcome the potentially exponential search space.

\section{Experimental evaluation}\label{sec:results}
We demonstrate the ability of \method{} to learn physically inspired
models with good extrapolation quality by experiments on synthetic
and real data.
For this, we implemented the network training and evaluation
procedure in \emph{python} based on the \emph{theano} framework~\cite{2016arXiv160502688short}.
We will make the code for training and evaluation public after acceptance of the manuscript. 

\paragraph{Pendulum.}
We first present the results of learning the equations of motion
for a very simple physical system: a pendulum.
The state space of a pendulum is $X=\Real\times\Real$ where the
first value is the angle of the pole in radians and the second
value is the angular velocity. In the physics literature, these
are usually denoted as $(\theta,\omega)$, but for our purposes,
we call them $(x_1,x_2)$ in order to keep the notation
consistent between experiments.
The pendulum's dynamic behavior is governed by the following
two ordinary differential equations:
\begin{equation}
\dot x_1  = x_2 \qquad\qquad\text{and}\qquad\qquad \dot x_2  = -g \sin(x_1)\,,\label{eqn:pend} 
\end{equation}
where $g=9.81$ is the gravitation constant.

We divide each equation by $g$ in order to balance the output scales
and form a regression problem with two output values, $y_1=\frac{1}{g}x_2$
and $y_2=-\sin(x_1)$.

As training data, we sample 1000 points uniformly in the hypercube
{\small $[-h,h] \times [-h,h]$} for $h=2$. Note that this domain contains more than
half of a sine period, so it should be sufficient to identify the analytic
expression.
The target values are disturbed by Gaussian noise with standard derivation
$\sigma=0.01$.
We also define three test sets, each with 1000 points.
The \emph{interpolation test set} is sampled from the same data distribution as the training set.
The \emph{extrapolation (near) test set} contains data sampled uniformly from the data domain {\small $[-\frac32 h,\frac32 h] \times [-\frac32 h,\frac32 h]\setminus [-h,h] \times [-h,h]$}, which is relatively near the training region and
 the \emph{extrapolation (far) test set} extends the region to further outside:
 {\small $[-2h,2h] \times [-2h,2h]\setminus [-h,h] \times [-h,h]$}.
We train a 2-layer \method{} and perform model selection among the hyper-parameters:
the regularization strength {\small $\lambda\in10^{\{-7,-6.3,-6,-5.3,-5,-4.3,-4,-3.3,-3\}}$}
and the number of nodes {\small $\frac 1 4 u=v\in\{1,3,5\}$}.
All weights are randomly initialized from a normal distribution with {\small $\sigma = \sqrt{1/(k'+d)}$}.
The unit selection $\typ{}$ is set such that all unit types are equally often.
To ensure convergence we chose $T=10000$ epochs.
We compare our algorithm to a standard multilayer
perceptron (MLP) with $\tanh$ activation functions and
possible hyperparameters: $\lambda$ as for \method,
number of layers {\small $L\in\{2,3\}$}, and number of neurons {\small $k\in\{5,10,20\}$}.
A second baseline is given by epsilon support vector regression (SVR)~\cite{basak2007:SVR} with
 two hyperparameters {\small $C\in10^{\{-3,-2,-1,0,1,2,3,3.5\}}$} and {\small $\epsilon \in 10^{\{-3,-2,-1,0\}}$}
 using radial basis function kernel with width {\small $\gamma\in \{0.05,0.1,0.2,0.5,1.0\}$}.

\begin{table}
  \caption{Numeric results on \emph{pendulum} dataset. Reported are the mean
    and standard deviation of the root mean squares error (RMS) ($\sqrt{E}$, \eqn{eqn:error}) on different test sets for 10 random initializations.
  }\label{tab:pend:results}
  \centering
  \begin{tabular}{l|l|l|l}
    \toprule
    & \text{interpolation} & \text{extrapol. (near)} & \text{extrapol. (far)}
    \\
    \text{\method{}} & $0.0102\pm 0.0000$ & $0.012\pm 0.002$ & $0.016\pm 0.007$ \\
    \text{MLP}     & $0.0138\pm 0.0002$ & $0.150\pm 0.012$ & $0.364\pm 0.036$ \\
    \text{SVR}     & $0.0105         $ & $0.041         $ & $0.18          $ \\
    \bottomrule
  \end{tabular}
\end{table}

\begin{figure}
  \centering
  \begin{tabular}{cc}
    (a)&(b)\\
    \includegraphics[height=0.22\linewidth]{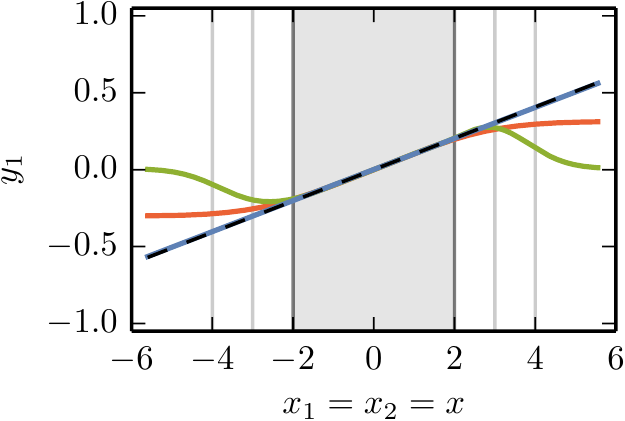}
    \includegraphics[height=0.22\linewidth]{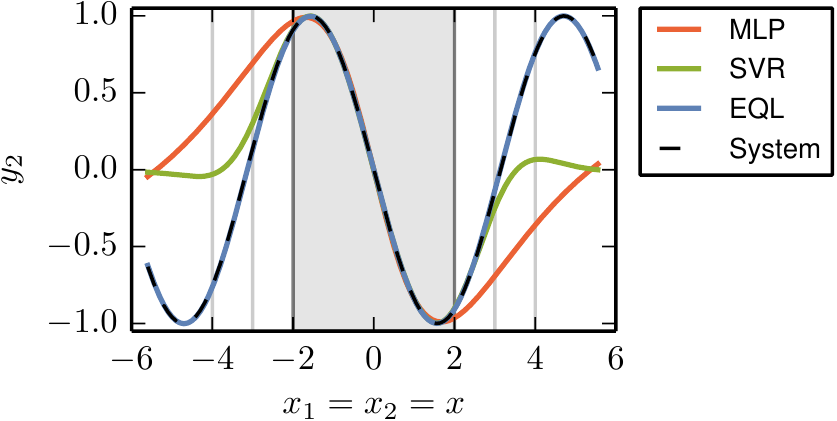}
    &\includegraphics[height=0.22\linewidth]{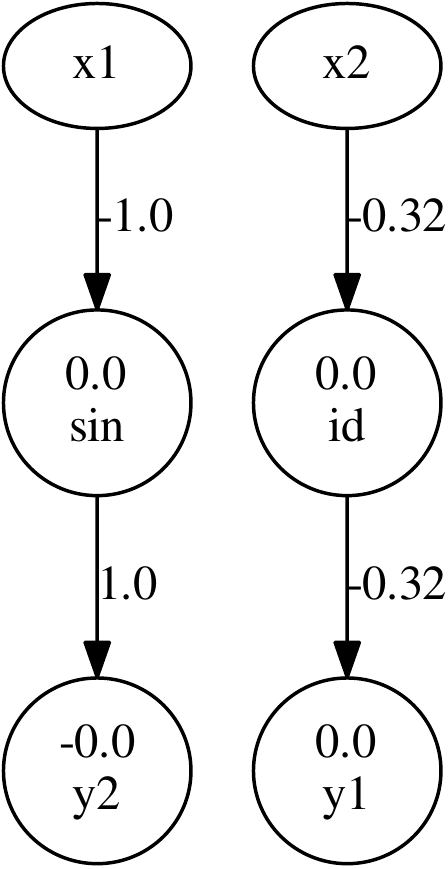}
  \end{tabular}\vspace{-.8em}
\caption{Learning pendulum dynamics.
  (a) slices of outputs $y_1$ (left) and $y_2$ (right) for inputs $x_1=x_2=x$ for the true system equation \eqnp{eqn:pend} and  one of \method{}, MLP, SVR instances. The shaded area marks the training region and the
  vertical bars show the size of the \emph{near} and \emph{far} extrapolation domain.
    (b) one of the learned networks. Numbers on the edges correspond to the entries of $W$ and numbers inside the nodes show the bias values $b$. All weights with $|w| < 0.01$ and orphan nodes are omitted. Learned formulas: $y_1=0.103 x_2$, $y_2=\sin(-x_1)$, which are correct up to symmetry ($\nicefrac{1}{g}=1.01$).}\label{fig:pend}
\end{figure}

 Numeric results are reported in \tab{tab:pend:results}.
As expected all models are able to interpolate well with a test error
on the order of the noise level ($\sigma=0.01$).
For extrapolation however, the performance differ between the
approaches.
For MLP the prediction quality decreases quickly when leaving the
training domain. SVR remains a bit better in the near extrapolation but also fails
 catastrophically on the far extrapolation data.
\method, on the other hand, extrapolates well,
both near and far away from the training domain.
The reasons can be seen in Figure~\ref{fig:pend}: while the MLP and SVR
simply learns a function that interpolates the training values,
\method{} finds the correct functional expression and
therefore predicts the correct values for any input data.

\begin{figure}
  \centering
  \begin{tabular}{@{}c@{~~}c@{~}c}
    (a)&(b)&(c)\\
     \raisebox{-.0\height}{\includegraphics[height=0.22\linewidth]{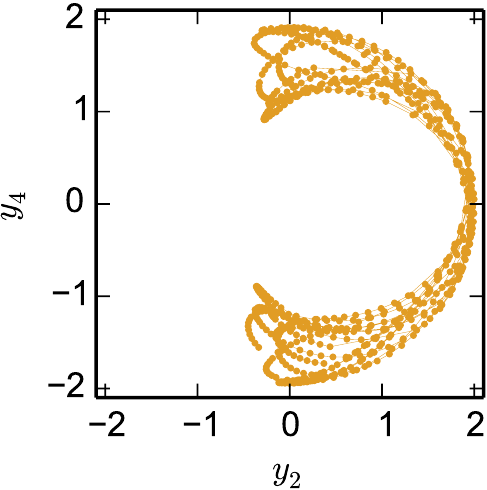}}
    &\raisebox{-.0\height}{\includegraphics[height=0.22\linewidth]{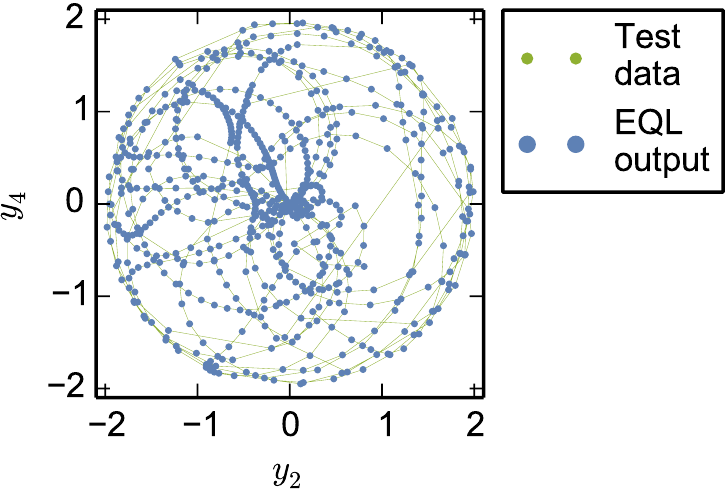}}
    &\raisebox{-.0\height}{\includegraphics[height=0.22\linewidth]{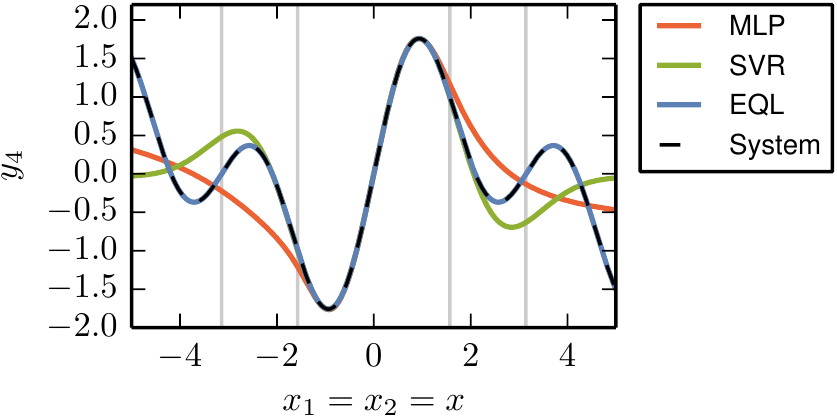}}
  \end{tabular}
  (d)~~{ \begin{tabular}{l|ccc}
                   & \method &  MLP & SVR  \\\hline
     \Tstrut
     extrapolation error & $0.0003\pm 0.00003$ & $0.58\pm 0.03$ & $0.25$\\
      \end{tabular}}
\caption{Double pendulum kinematics.
  (a)~training trajectory (in y-space).
  (b)~extrapolation test trajectory (in y-space) with output of a learned \method{} instance.
  (c)~slices of output $y_4$ for inputs $x_1=x_2=x$ for the true system, one of \method{}, MLP, and SVR instances.
  (d)~numeric results, see~\tab{tab:pend:results} for details. Note, that predicting 0 would yield a mean error of $0.84$.
  }\label{fig:dpk}
\end{figure}

\paragraph{Double pendulum kinematics.}
The second system we consider real double pendulum where the forward kinematics should be learned.
For that we use recorded trajectories of a real double pendulum \cite{SchmidtLipson2009:learnnaturallaws}.
The task here is to learn the position of the tips of the double pendulum segments
 from the given joint angles ($x_1,x_2$). These positions where not measured such that we supply them
 by the following formula:
$y_1=\cos(x_1), y_2=\cos(x_1)+\cos(x_1+x_2), y_3=\sin(x_1), y_4=\sin(x_1)+\sin(x_1+x_2)$
where
 $(y_1,y_3)$ and $(y_2,y_4)$ correspond to x-y-coordinates of the first and
 second end-point respectively.
The dataset contains two short trajectories.
The first covers only part of the domain (input as well as output)
 and consists of 819 samples where 10\% was used as validation set (randomly sampled), see \fig{fig:dpk}(a).
The second trajectory corresponds to a behavior with several spins of both pendulum segments such
 that a much larger domain is covered. Nevertheless the angle values are confined to $[-\pi,\pi]$.
We use this trajectory as extrapolation test set.
The trajectory and the outputs of our method are shown in \fig{fig:dpk}(b).
The prediction for unseen domains is perfect,
 which is also illustrated in a systematic sweep, see \fig{fig:dpk}(c).
The performance of MLP is off already near the training domain. SVR is a bit better, but still
 does not give usable predictions for the test data, see also the root means square error in \fig{fig:dpk}(d).

Model selection is performed to determine $\lambda$ as above,
 $u=v\in\{3,5\}$, (MLP: $k\in\{5,10,20\}$) and layer number $L\in\{2,3\}$.

\paragraph{Robotic arms.}
A more complicated task is to learn the forward kinematics of multi-segment robotic arms.
We consider planar arms with 3, 4, and 5 joints, where each segment is 0.5 units long.
For training the arm is controlled by sinusoidal joint target angles with amplitude in $[-\nicefrac{\pi}{2},\nicefrac{\pi}{2}]$, each joint with a different frequency. The number of data points are: 3000, 6000, and 18000 for the 3, 4, and 5 segment arms respectively, with added noise as above.
For testing extrapolation performance the amplitude $[-\pi,\pi]$ was used. Note that the extrapolation
 space is much larger than the training space.
The task is to predict the coordinates of the end-effector of the arms (\emph{kin-3-end}, \emph{kin-4-end})
 and the coordinates of all segment positions \emph{kin-5-all}.
The numerical results, see \tab{tab:kin}, shows that our method is able to extrapolate in these cases.
Model selection as above with $u=v\in\{10,20\}$, (MLP: $k\in\{10,50\}$) and layer number $L\in\{2,3,4\}$.
To illustrate the dependence on the amount of noise and the number of available training points we provide a quantification in Appendix A2. In short, increasing noise can be compensated by increasing amount of data to keep the performance. 

\begin{table}
  \caption{Extrapolation performance for \emph{kinematic of robotic arms}.
    See \tab{tab:pend:results} for details. Standard deviations for 5 random initializations.
    Interpolation error for all methods is around $0.012 \pm 0.02$}
  \label{tab:kin}
  \centering
  \begin{tabular}{l|llll}
    \toprule
                & kin-3-end          & kin-4-end      & kin-5-all  \\
                \hline\Tstrut
    \text{EQL}  & $0.028\pm 0.019$  & $0.018\pm 0.004$ & $0.036\pm 0.035$\\
    \text{MLP}  & $0.369\pm 0.041 $  & $0.415\pm 0.020$ & $0.346\pm 0.013$\\
    \text{SVR}  & $0.235          $  & $0.590         $ & $0.260         $\\
    \bottomrule
  \end{tabular}
\end{table}

\paragraph{Learning complex formula.}
In order to find out whether \method{} can also learn more complicated
formulas, we consider three examples with four-dimensional input and
one-dimensional output:
\begin{align}
y &= \nicefrac{1}{3} \left(\sin(\pi x_1) + \sin\left(2 \pi x_2 + \nicefrac{\pi}{8}\right)+x_2 - x_3 x_4 \right)&\text{F-1}\label{eqn:syn1}\\
y &= \nicefrac{1}{3} \left(\sin(\pi x_1) + x_2 \cos(2\pi x_1 + \nicefrac{\pi}{4}) + x_3-x_4^2\right) &\text{F-2}\label{eqn:syn2}\\
y &= \nicefrac{1}{3} \left( (1+x_2)  \sin(\pi x_1) + x_2  x_3  x_4\right) &\text{F-3}\label{eqn:syn3}
\end{align}
The first equation requires only one hidden layer to be represented.
The second equation and third equation should requires two hidden layers.
In particular, F-2 contains a product of $x_2$ and $\cos$ and F-3 contains
a product of three terms, and we use it to test if our restriction
to only pairwise product units causes problems for more complex
target functions.
We follow the same procedure as in the pendulum case for building
training and test sets, though with $h=1$ as input data range.
We use 10000 points for training set and validation set (90\%-10\% split)
and 5000 points for each of the test sets.
Model selection for \method{} is performed as above
 using the number of layers {\small $L\in{2,3,4}$}.
 The number of units is set to $\frac{1}{4}u=v=10$.
For the MLP, we select $L$ and $\lambda$ from the same set as above as well as {\small $k\in\{10,30\}$}.

\begin{figure}
  \centering
  \begin{tabular}{cc}
    \multicolumn{2}{c}{(a) F-1}\\
    \includegraphics[height=0.22\linewidth]{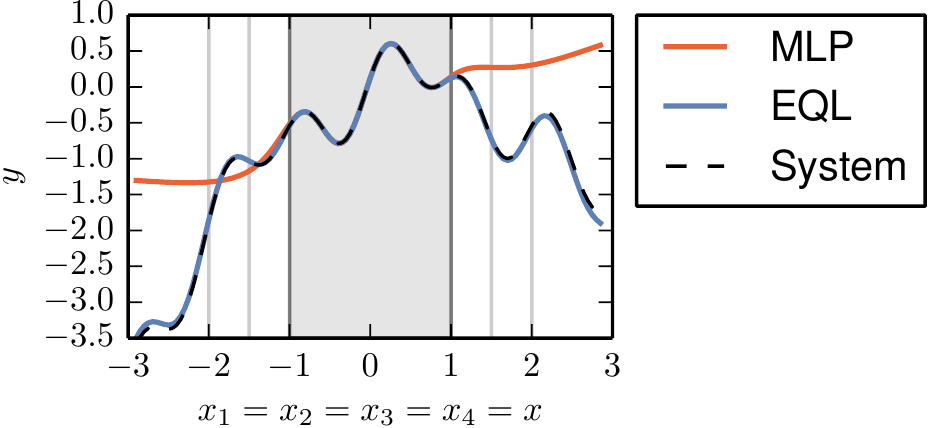}
    &\includegraphics[height=0.22\linewidth]{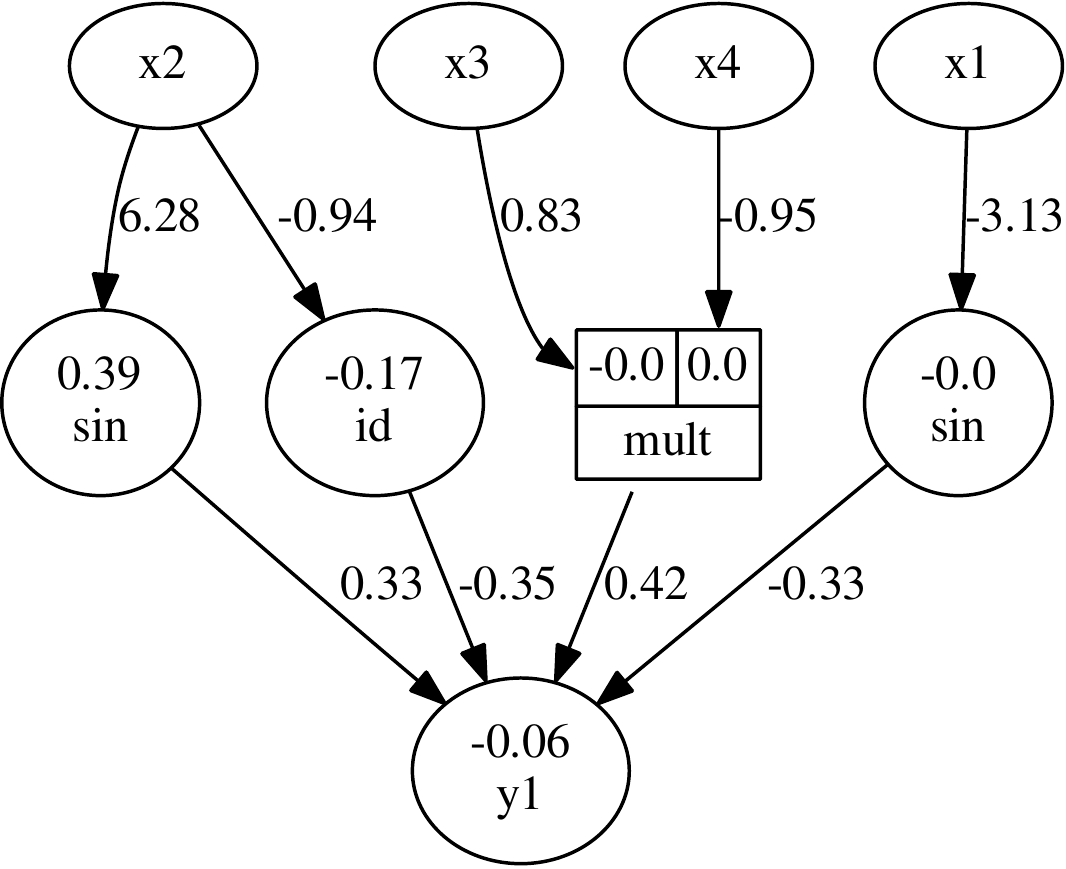}\\
    \multicolumn{2}{l}{learned formula: \scriptsize $-0.33 \sin(-3.13 x_1) + 0.33 \sin( 6.28 x_2 + 0.39) + 0.33 x_2 -0.056 - 0.33 x_3 x_4$}\\[.2em]
    \multicolumn{2}{c}{(b) F-2}\\
    \includegraphics[height=0.22\linewidth]{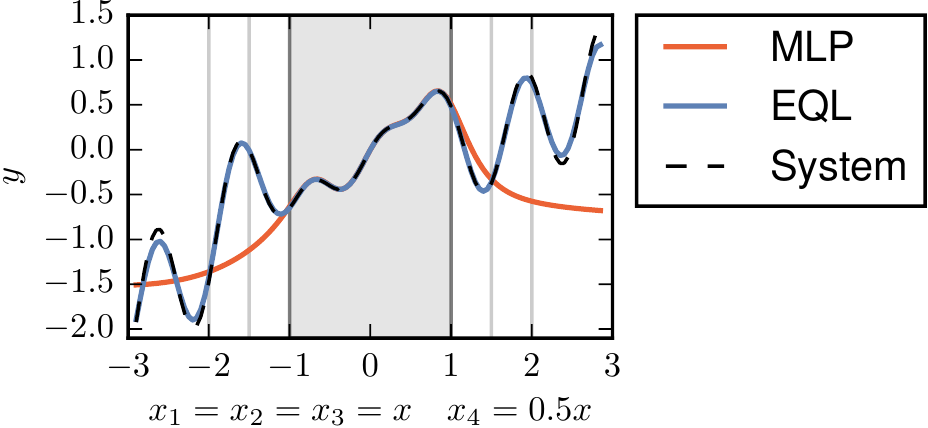}
    &\includegraphics[height=0.22\linewidth]{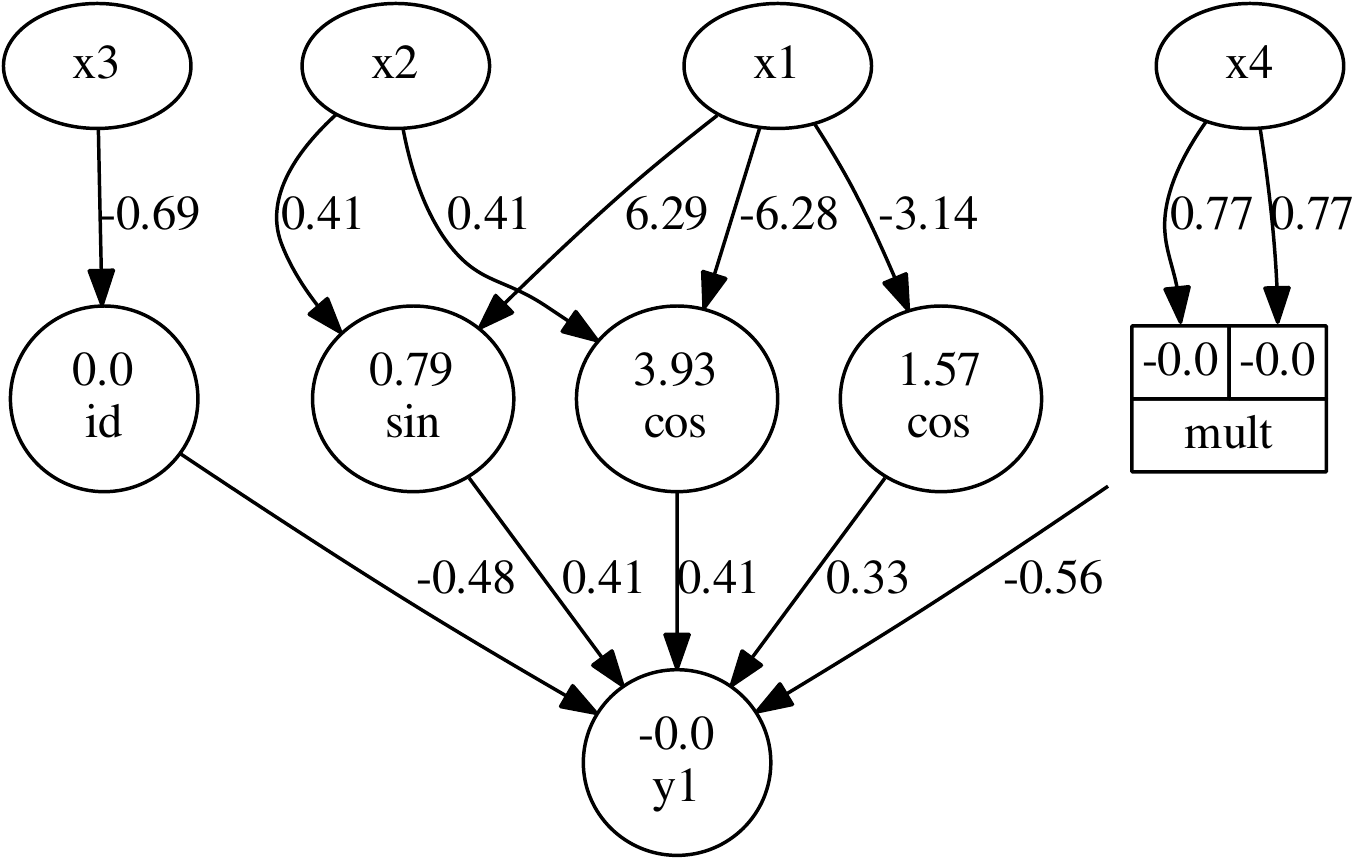}\\
    \multicolumn{2}{l}{learned formula: \scriptsize
      $0.33 \cos(3.14 x_1+1.57) + 0.33 x_3 - 0.33 x4^2 + $}\\
    \multicolumn{2}{r}{\scriptsize $0.41 \cos(-6.28 x_1 + 3.93 + 0.41 x_2) + 0.41 \sin(6.29 x_1+0.79+0.41 x_2)$}\\[.2em]
    \multicolumn{2}{c}{(c) F-3}\\
    \includegraphics[height=0.22\linewidth]{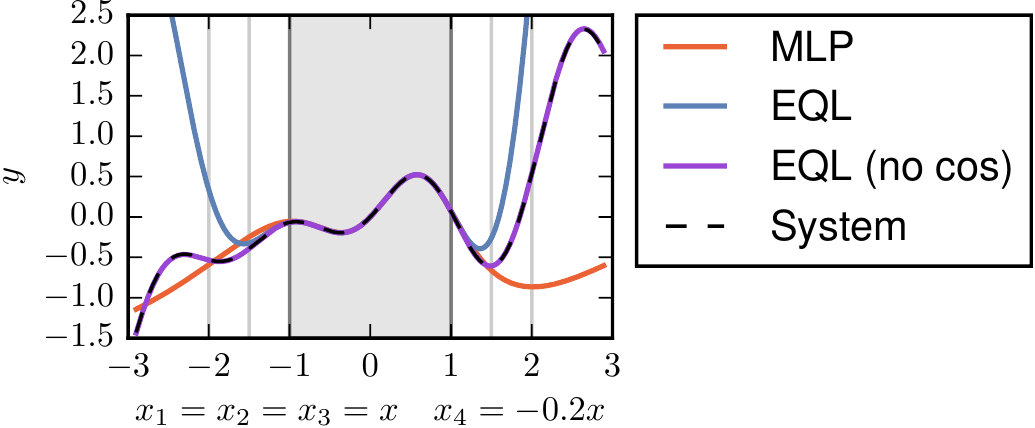}
    & \includegraphics[height=0.22\linewidth]{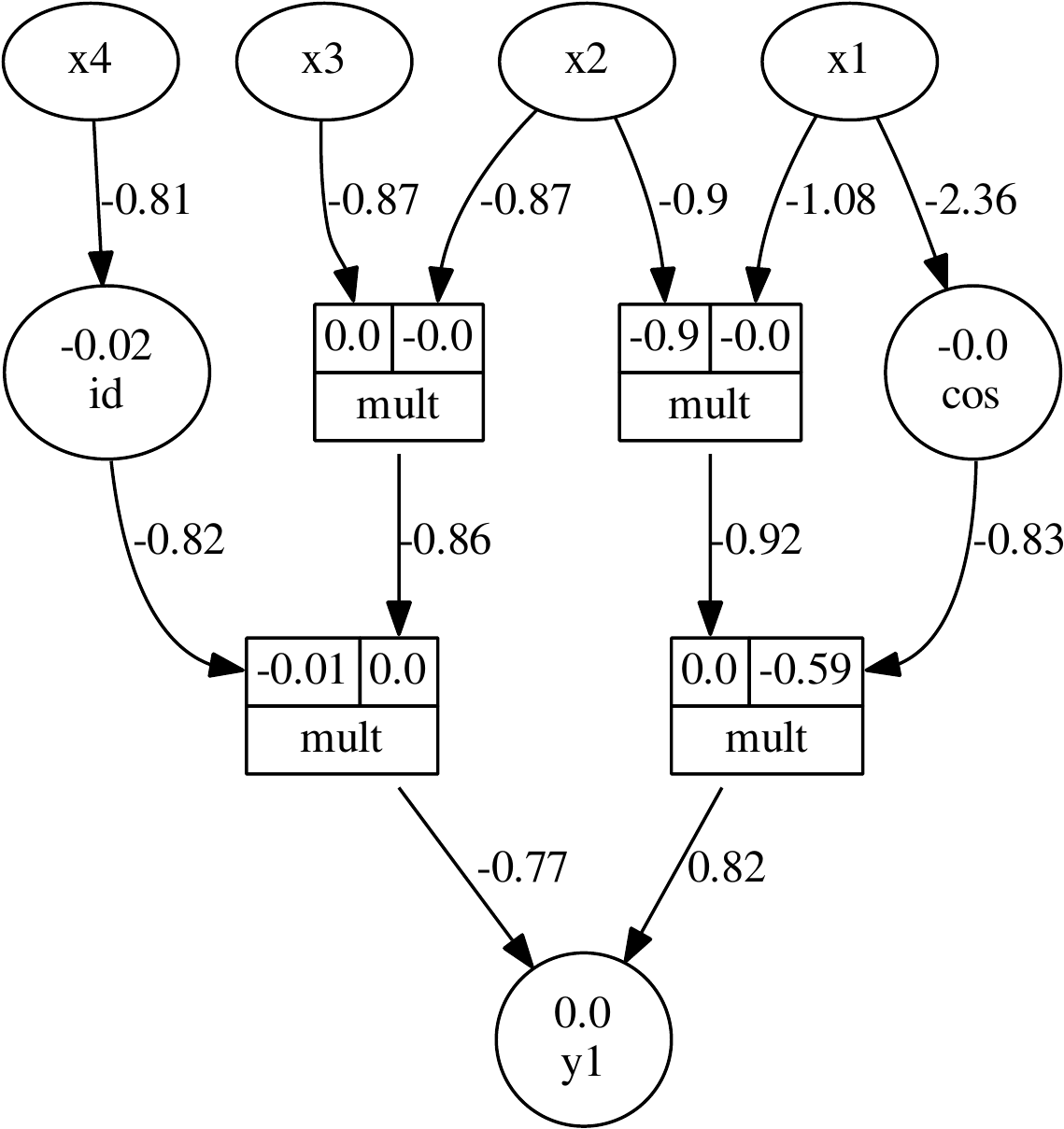}\quad
    \includegraphics[height=0.22\linewidth]{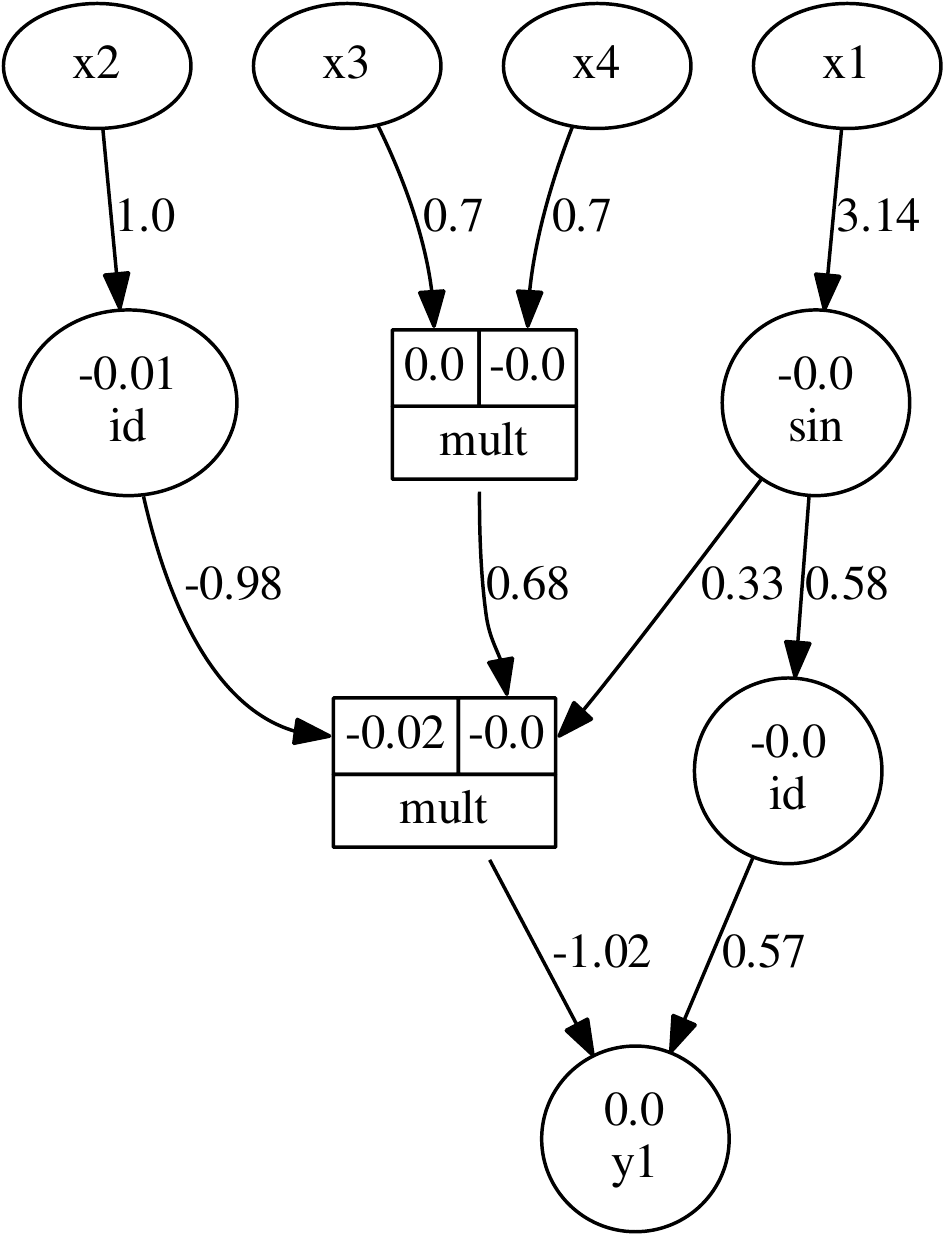}\\[-1.2em]
     & \hspace{0.03\linewidth} EQL \hspace{0.09\linewidth} EQL (no cos)\hfill \ \\[0.2em]
    \multicolumn{2}{l}{learned formula (EQL): \scriptsize
      $0.61(x_1 + x_1x_2)(\cos(-2.36 x_1)+0.71)  +  0.33 x_2 x_3 x_4$}\\
    \multicolumn{2}{l}{learned formula (EQL (no cos)): \scriptsize
      $0.33 (1+x_2)\sin(3.14 x_1) +  0.33 x_2 x_3 x_4$}\\
      \end{tabular}
\caption{Formula learning analysis. (a) for F-1, (b) for F-2, and (c) for F-3.
  (left) $y$ for a single cut through the input space for the true system equation (\ref{eqn:syn1},\ref{eqn:syn2}), and for an instance of \method{}, and MLP.
   (right) shows the learned networks correspondingly, see \fig{fig:pend} for details.
  The formula representations where extracted from the networks.
  For F-3 the algorithm fails with the overcomplete base and typically (9/10 times)
  ends up in a local minima.
  With less base function (no cosine) the right formula is found. Both results are presented. See text for a discussion.
}\label{fig:syn}
\end{figure}

\Tab{tab:syn:results} shows the numerical results.
Again, all methods are able to interpolate, but only \method{} achieves
good extrapolation results, except for equation F-3. There it settles in 9 out of 10 cases
 into a local minimum and finds only an approximating equation that deviates outside the training domain.
Interestingly, if we restrict the base functions to not contain cosine, the algorithm finds the right formula. Note, the sparsity of the correct formula is lower than those of the approximation, so it should be selected if found.
Figure~\fig{fig:syn} illustrates the performance and the learned networks visually.
It shows one of the model-selected instances for each case.
For F-1 the correct formula was identified,
so correct predictions can be made even far outside the training region (much further than illustrated).
For F-2 the network provided us with a surprise, because it yields good extrapolation performance with only one hidden layer! How can it implement $x_2\cos(a x_1+b)$? Apparently it uses $1.21 \cos(a x_1 + \pi + b + 0.41 x_2) + \sin(a x_1 + b + 0.41 x_2)$ which is a good approximation for $x_2 \in [-2,2]$.
The sparsity of this solution is $5$ whereas the true solution needs at least $6$, which explains its selection.
For F-3 the suboptimal local minima uses some strange way of approximating $(1+x_2)\sin(x_1)$ using $(x_1 + x_1 x_2)\cos(\beta x_1)$, which deviates fast, however the true solution would be sparser but was not found. Only if we remove cosine from the base functions we get always the correct formula, see \fig{fig:syn}(c).

\begin{table}
  \caption{Interpolation and extrapolation performance for \emph{formula learning}. See \tab{tab:pend:results} for details.
  }\label{tab:syn:results}
  \centering
  \begin{tabular}{ll|lll}
    \toprule
    dataset&method& \text{interpolation} & \text{extrapol. (near)} & \text{extrapol. (far)}
    \\ \hline
    F-1\Tstrut
    &\method    & $0.010\pm 0.000$ & $0.015\pm 0.005$ & $0.026\pm 0.015$ \\
    &\text{MLP} & $0.011\pm 0.000$ & $0.32\pm 0.12$   & $0.920\pm 0.420$ \\
    &\text{SVR} & $0.011$          & $0.28$           & $1.2$ \\
    \hline
    F-2\Tstrut
    & \method    & $0.01\pm 0.00 $ & $0.013\pm 0.004$  & $0.026\pm 0.019$ \\
    & \text{MLP} & $0.01\pm 0.00 $ & $0.2\pm 0.014   $ & $0.49\pm 0.043 $ \\
    & \text{SVR} & $0.011        $ & $0.3            $ & $ 0.94         $  \\
    \hline
    F-3\Tstrut
    &\method    & $0.01\pm 0.000$  & $0.047\pm 0.012$  & $0.35\pm 0.11$ \\
    &\method\  (no cos) & $0.01\pm 0.000$ & $0.01\pm 0.000$ & $0.011\pm 0.001$ \\
    &\text{MLP}  & $0.01\pm 0.000$ & $0.084\pm 0.007$  & $0.4\pm 0.021$ \\
    &\text{SVR}  & $0.01$          & $0.071          $ & $0.39$ \\
    \bottomrule
  \end{tabular}
\end{table}

\begin{figure}
  \centering
  \begin{tabular}{ccc@{}l}
    (a)&(b)&(c)\\
    \includegraphics[height=0.18\linewidth]{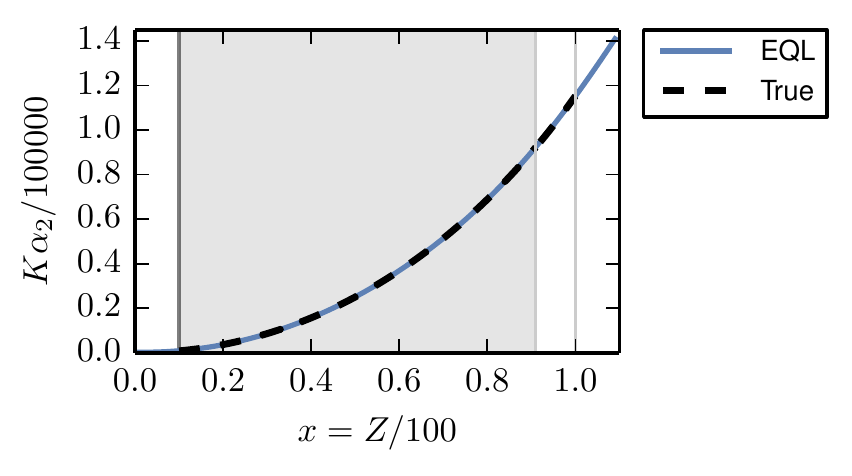}&
    \includegraphics[height=0.18\linewidth]{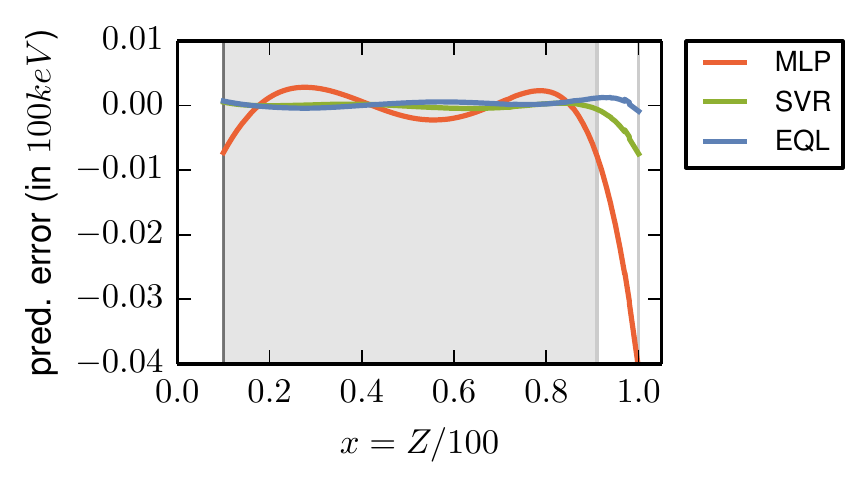}&
    \includegraphics[height=0.16\linewidth]{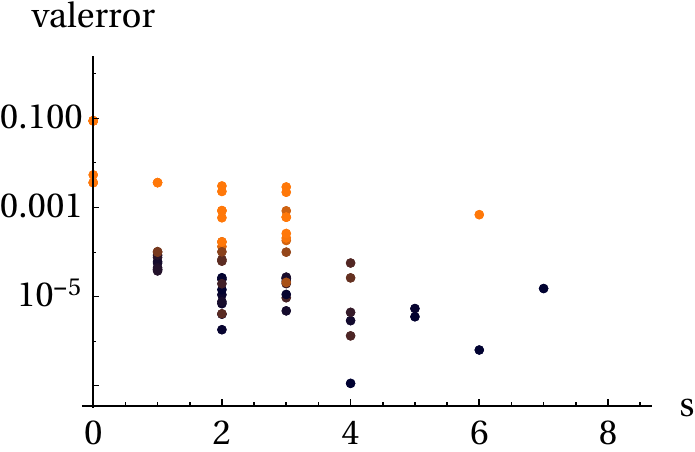}&
    \hspace*{-10pt}\includegraphics[height=0.18\linewidth, trim=0 -10pt 0 0]{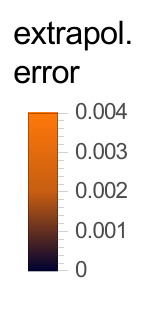}\\
  \end{tabular}

  \begin{minipage}{.5\linewidth}\centering
    (d)\\
  \begin{tabular}{l|ll}
    \toprule
              & interpolation      & extrapolation    \\
              \hline\Tstrut
 \method      & $0.00042  $& $0.0061\pm 0.0038 $\\
 \text{MLP}   & $0.002    $& $0.0180\pm 0.0024 $\\
 \text{SVR}   & $0.00067  $& $0.0057\pm 0.0014 $\\
    \bottomrule
  \end{tabular}
\end{minipage}
  \begin{minipage}{.45\linewidth}\centering
    (e)\\
    \scriptsize
    \begin{tabular}{l|l}
      s & formula\\
      \hline
      \Tstrut
      1 &  $y = 1.28 x^2 - 0.183 x + 0.026$\\ 
      2 &  $y = 1.98 x^2 - 1.42 x + 0.618 - 1.45 \sigm(-3.65 x - 0.3)$\\ 
      3 & $y = -0.38 z + 2.47 \sigm(-2.25 z - 2.77) + 0.38$\\
        & \quad with \quad $z = \cos(2.32 x - 0.08)$\\  
      4 & $y = 0.221 z + 0.42 \sigm(0.75 z - 3.73)$\\ 
        & \quad with \quad $z = 4.65 x^2 - 0.229 x$
    \end{tabular}
  \end{minipage}
  \vspace{-.5em}
  \caption{X-Ray transition energies.
    (a) Measured data and predicted values by \method{} and
    (b) visualized prediction error for all methods for one train/validation splitting.
    (c) \method{} solutions during model selection in validation error --  sparsity space,
      see Appendix A1 for details.
    (d) numeric results. Reported are RMS errors with standard deviation for 10 independent train/validation splits. In real units the error is in 100\,keV and is well below the difference between neighboring high-$Z$ elements.
    (e) learned formulas for different sparsities $s$ (lowest dot for each $s$ in (c)).
  }
  \label{fig:xray}
\end{figure}

\paragraph{X-Ray transition energies.}
As a further example we consider data measured in atomic physics.
When shooting electron beams onto atoms one can excite them and they consequently emit
 x-ray radiation with characteristic peak energies.
 For each element/isotope these energies are different as they correspond to the
 potential difference between the electron shells, such that one can identify elements in a probe
  this way.
The data is taken from~\cite{Deslattes2003:XrayTransEnergies},
 where we consider one specific transition, called the $K\,\alpha_2$ line, because it was measured for all elements.
The true relationship between atomic number $Z$ and transition energies is complicated, as it
 involves many body interactions and no closed-form solution exists.
Nevertheless we can find out which relationships our system proposes.
It is known that the main relationship is $K\,\alpha_2 \propto Z^2$ according to Moseley's law. Further correction
 terms for elements with larger $Z$ are potentially of higher order.
We have data for elements with $10\le Z \le 100$, which is split into training/validation sets in the range $[10,91]$ (70/10 data points)
and extrapolation test set in the interval $[92,100]$ (14 data points because of isotops).
Since we have so little data we evaluate the performance for 10 independent training/validation splits.
The data is scaled to lie in $[0,1]$, \ie $x= Z/100$ and $y=K\alpha_2/100000$.
Model selection is here based on validation error only. The selection for sparsity and validation error only yields the $Z^2$ relationship. Mini-batch size is 2 here and $T=50000$ was used.
\Fig{fig:xray} presents the data, the predictions, the learned formulas and the numerical results.
\method{} and SVR achieve similar performance and MLP is significantly worse.
However, \method{} also yields interpretable formulas,
see \fig{fig:xray}(e) that can be used to gain insights into the potential relationship.

\subsection{Poor extrapolation out of model class --- cart-pendulum system}
Let us now go beyond our assumptions and consider cases where
the true target function is not an element of the hypothesis set.

\begin{figure}
  \centering
  \begin{tabular}{ccc}
    (a)&(b)&(c)\\
    \includegraphics[height=0.18\linewidth]{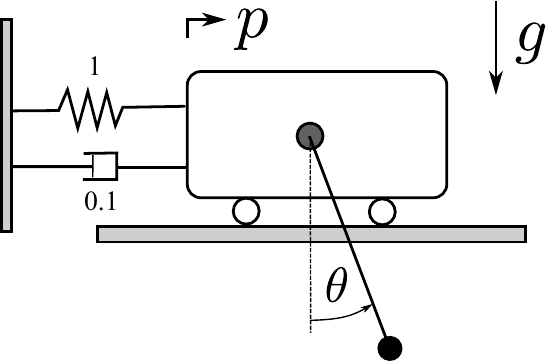}
    &\includegraphics[height=0.18\linewidth]{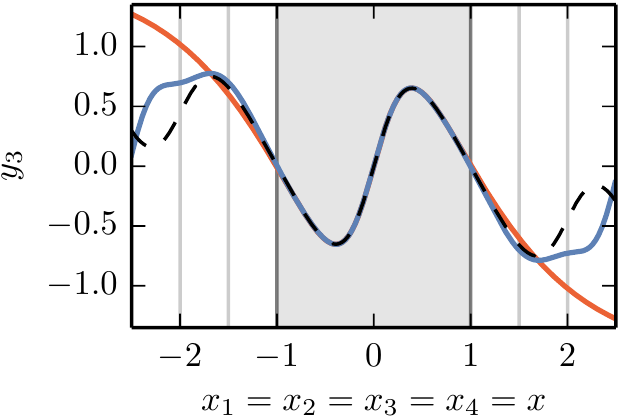}
    &\includegraphics[height=0.18\linewidth]{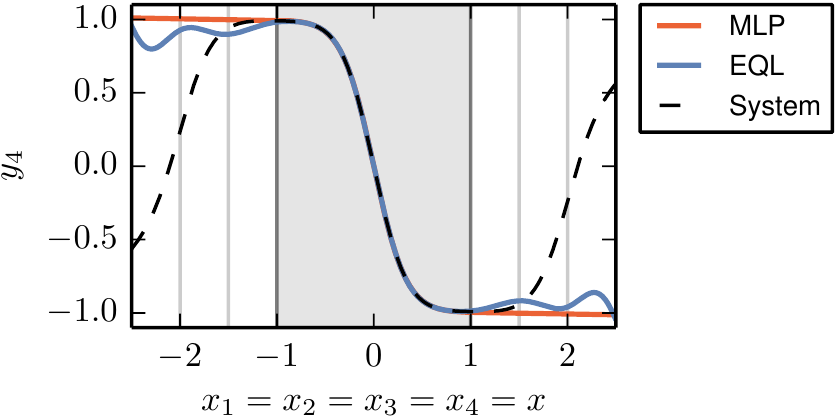}
  \end{tabular}
  \caption{Cart-pendulum system. (a) sketch of the system.
     The lengths and masses are set to 1, the gravitation constant is $9.81$ and the friction constant is $0.01$.
    (b,c) slices of outputs $y_3$ and $y_4$ for inputs $x_1=x_2=x_3=x_4=x$
     for the true system equation \eqnp{eqn:cp}, and best \method{}, MLP instances.
   }
  \label{fig:cp}
\end{figure}

Consider a pendulum attached to a cart
that can move horizontally along a rail but that is attached to a
spring damper system, see \fig{fig:cp}(a).
The system is parametrized by 4 unknowns: the position of the cart,
the velocity of the cart, the angle of the pendulum and the angular
velocity of the pendulum. We combine these into a four-dimensional
vector $x=(x_1,\dots,x_4)$.

We set up a regression problem with four outputs from the corresponding
system of ordinary differential equations where $y_1 = \dot x_1 = x_3$, $y_2 = \dot x_2 = x_4$ and
\begin{align}
  y_3&= \frac{-x_1-0.01 x_3+x_4^2 \sin\left(x_2\right)+0.1 x_4
    \cos \left(x_2\right)+9.81 \sin \left(x_2\right) \cos
    \left(x_2\right)}{\sin ^2\left(x_2\right)+1}\label{eqn:cp}, \\
  y_4&= \frac{-0.2 x_4 - 19.62 \sin \left(x_2\right) + x_1
    \cos \left(x_2\right) + 0.01 x_3  \cos \left(x_2\right) - x_4^2
    \sin \left(x_2\right)\cos \left(x_2\right)}
  {\sin^2\left(x_2\right)+1}.\nonumber
\end{align}
The formulas contain divisions which are not included in our architecture due to their
 singularities. To incorporate them in a principled manner is left for future work.
Thus, the cart-pendulum dynamics is outside the hypothesis class.
In this case we {\bf cannot} expect great extrapolation performance
 and this is confirmed by the experiments.
In \fig{fig:cp}(b,c) the extrapolation performance is illustrated by slicing through the input space.
The near extrapolation performance is still acceptable for both \method{} and MLP,
 but as soon as the training region is left further even
 the best instances differ considerably from the true values, see also the numeric results in \tab{tab:cp:results}. The SVR is performing poorly also for near extrapolation range.
Inspecting the learned expressions we find that the sigmoid functions are rarely used.

\begin{table}
\caption{Interpolation and extrapolation performance for \emph{cart-pendulum dynamics}.
  See \tab{tab:pend:results} for details. Note that predicting 0 would yield an error of 0.96 on the far test set.}\label{tab:cp:results}
\centering
\begin{tabular}{l|lll}
  \toprule
   & \text{interpolation} & \text{extrapol. (near)} & \text{extrapol. (far)} \\
  \hline \Tstrut
  \method      & $0.0103\pm 0.0000$ & $0.0621\pm 0.0208$ & $0.180\pm 0.056$ \\
  \text{MLP}   & $0.0101\pm 0.0000$ & $0.0184\pm 0.0008$ & $0.195\pm 0.006$ \\
  \text{SVR}   & $0.0118          $ & $0.227           $ & $0.639         $ \\
 \bottomrule
\end{tabular}
\end{table}

\vspace*{-.2em}
\section{Conclusions}\vspace*{-.2em}
We presented a new network architecture called \method{}
 that can learn analytic expressions that typically occur in
 equations governing physical, in particular mechanical, systems.
The network is fully differentiable, which allows end-to-end training
using backpropagation.
By sequencing $L_1$ regularization and fixing $L_0$ norm
 we achieve sparse representations with unbiased estimation of factors within the learned equations.
We also introduce a model selection procedure
 specifically designed to select for good extrapolation quality
 by a multiobjective criterion based on validation error and sparsity.
The proposed method is able to learn
functional relations and extrapolate them to unseen parts of the
data space, as we demonstrate by experiments on synthetic as well
as real data.
The approach learns concise functional forms that may provide insights into
the relationships within the data, as we show on physical measurements of x-ray transition energies.

The optimization problem is nontrivial and has many local minima.
We have shown cases where the algorithm is not reliably finding the right equation
 but instead finds an approximation only, in which case extrapolation may be poor.

If the origin of the data is not in the hypothesis class,
\ie the underlying expression cannot be represented by the network,
 good extrapolation performance cannot be achieved.
Thus it is important to increase the model class by incorporating more base functions
 which we will address in future work alongside the
 application to larger examples.
We expect good scaling capabilities to larger systems
 due to the gradient based optimization.
Apart from the extrapolation we also expect improved interpolation results
 in high-dimensional spaces, where data is less dense.

\subsubsection*{Acknowledgments}
GM received funding from the People Programme (Marie Curie Actions) of the European Union's Seventh Framework Programme (FP7/2007-2013) under REA grant agreement no.~[291734].

\bibliography{biblioML}

\appendix
\section{Appendix}
\section{A1: Model selection details}\label{sec:modelsel:app}
\subsection{Quantifying sparsity}
We actually want a measure of complexity of the formula, however, since it is not clear what is the right
 choice of a measure, we use the sparsity instead, by counting the number of active/used hidden units
 denoted by $s$. For a given network $phi$ we get
 \begin{align}
   s(\phi) = \sum_{l=1}^L\sum_{i=1}^k\Theta( |W^\l_{i,\cdot}| * |W^{\layer{l+1}}_{\cdot,i}| - 0.01)\,,\label{eqn:s}
 \end{align}
where $\Theta$ is the heavyside function and 0.01 is an arbitrary threshold. For the multiplication units the norm of the incoming weights for both inputs are added (omitted to avoid clutter in the formula).

\subsection{Selection criteria}
\begin{figure}[bhp]
  \centering
  \includegraphics[width=.6\linewidth]{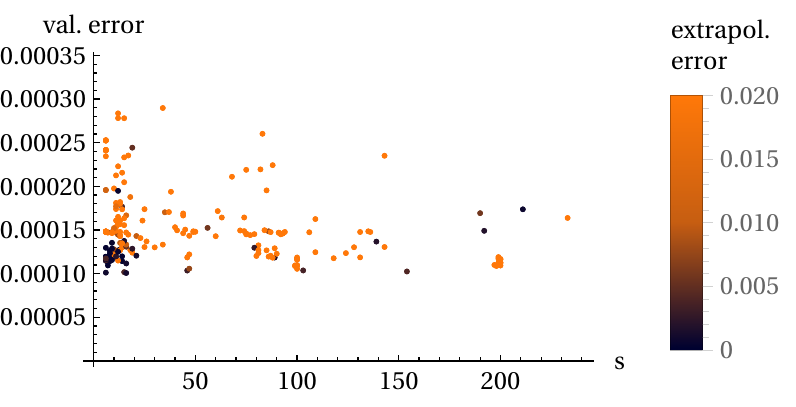}
  \caption{Extrapolation performance depending on validation error and sparsity ($s$) for the kin2D-3 dataset as an illustration.\label{fig:model:sel}
  }
\end{figure}

As stated in the main text, we strive to choose the model that is both simple and has good performance in terms
 of the validation set.
Since both quantities have different scales, we proposed to choose them based on their ranking.
Let $r^v(\phi)$ and $r^s(\phi)$ be the ranks of the network $\phi$ \wrt the validation error and sparsity $s(\phi)$respectively, then
 the network with minimal squared rank norm is selected:
 \begin{align}
   \argmin_\phi\left[ r^v(\phi)^2 + r^s(\phi)^2\right] \label{eqn:model:sel}
 \end{align}
In \fig{fig:model:sel} the extrapolation performance of all considered networks for the kin2D-3 dataset is visualized in dependence of validation error and the sparsity. It becomes evident that the best performing networks are both sparse and have a low validation error.

\section{A2: Dependence on noise and number of data points}\label{sec:dep:noise-pts}
\begin{figure*}
  \centering
  \begin{tabular}{cc}
    (a)&(b)\\
    \includegraphics[width=.48\linewidth]{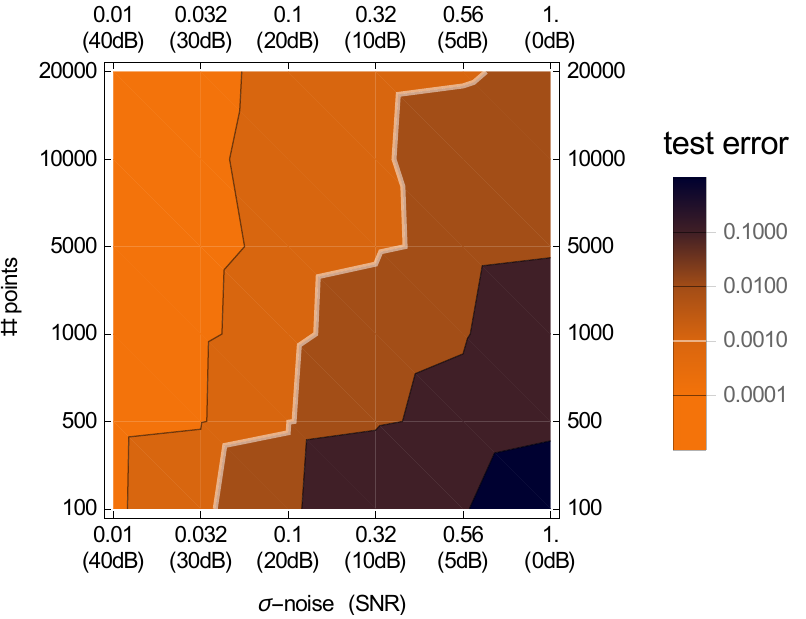}&
    \includegraphics[width=.48\linewidth]{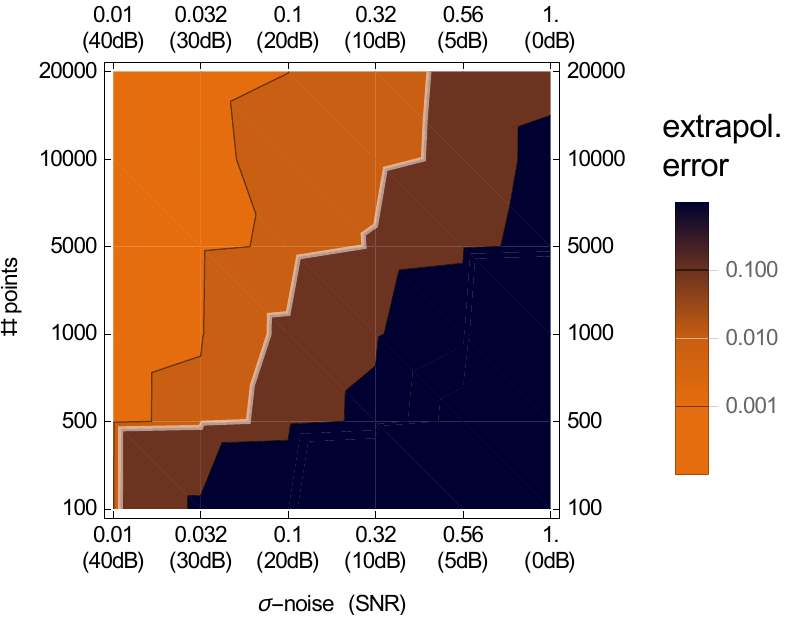}
  \end{tabular}
  \caption{Interpolation performance (a) and extrapolation performance (b) (on the noise-free test set) depending on the number of data points and the size of the additive noise for \emph{kin-4-end} dataset as an illustration.
    The white line represent an arbitrary threshold below which we consider a successful solution of the interpolation and extrapolation task.
    \label{fig:dep:noise-pts}
  }
\end{figure*}

In order to understand how the method depends on the amount of noise and the number of datapoints
 we scan through the two parameters and present the empirical results in \fig{fig:dep:noise-pts}.
In general the method is robust to noise and as expected, more noise can be compensated by more data.

\end{document}